\documentclass[sigconf]{acmart}
\renewcommand\footnotetextcopyrightpermission[1]{}
\AtBeginDocument{%
  }

\setcopyright{acmlicensed}
\copyrightyear{2025}
\acmYear{2025}
\acmConference[ACM Multimedia]{}{October 27--31,
  2025}{Dublin, Ireland}

\usepackage{booktabs}
\usepackage{multirow}
\usepackage{pifont}         
\newcommand{\cmark}{\ding{51}} 
\newcommand{\xmark}{\ding{55}} 
\usepackage{graphicx}
\usepackage{subcaption}
\usepackage{hyperref}
\begin{document}

\title{ReSeDis: A Dataset for Referring-based Object Search across Large-Scale Image Collections}

\author{Ziling Huang}
\email{huangziling@nii.ac.jp}
\affiliation{%
  \institution{National Institute of Informatics}
  \city{Tokyo}
  \country{Japan}
}

\author{Yidan Zhang}
\email{zhang-yidan@g.ecc.u-tokyo.ac.jp}
\affiliation{%
  \institution{The University of Tokyo}
  \city{Tokyo}
  \country{Japan}
}

\author{Shin'ichi Satoh}
\email{satoh@nii.ac.jp}
\affiliation{%
  \institution{National Institute of Informatics}
  \city{Tokyo}
  \country{Japan}
}

\renewcommand{\shortauthors}{Ziling et al.}

\begin{abstract}
Large‐scale visual search engines are expected to solve a dual problem at once: (i) locate every image that truly contains the object described by a sentence and (ii) identify the object’s bounding box or exact pixels within each hit. Existing techniques address only one side of this challenge. Visual Grounding yields tight boxes and masks but rests on the unrealistic assumption that the object is present in every test image, producing a flood of false alarms when applied to web-scale collections. Text-to-image retrieval excels at sifting through massive databases to rank relevant images, yet it stops at whole-image matches and offers no fine-grained localization.

We introduce Referring Search and Discovery (ReSeDis), the first task that unifies corpus-level retrieval with pixel-level grounding. Given a free-form description, a ReSeDis model must decide whether the queried object appears in each image and, if so, where it is, returning bounding boxes or segmentation masks. To enable rigorous study, we curate a benchmark in which every description maps uniquely to object instances scattered across a large, diverse corpus, eliminating unintended matches. We further design a task-specific metric that jointly scores retrieval recall and localization precision. Finally, we provide a straightforward zero-shot baseline using a frozen vision–language model, revealing significant headroom for future study. ReSeDis offers a realistic, end-to-end testbed for building the next generation of robust and scalable multimodal search systems. The dataset is available at \href{https://github.com/hufflepuff0596/ReSeDis}{https://github.com/hufflepuff0596/ReSeDis}.
\end{abstract}



\maketitle

\section{Introduction}
\begin{figure*}[ht]

\centering
\includegraphics[width=0.75\linewidth]{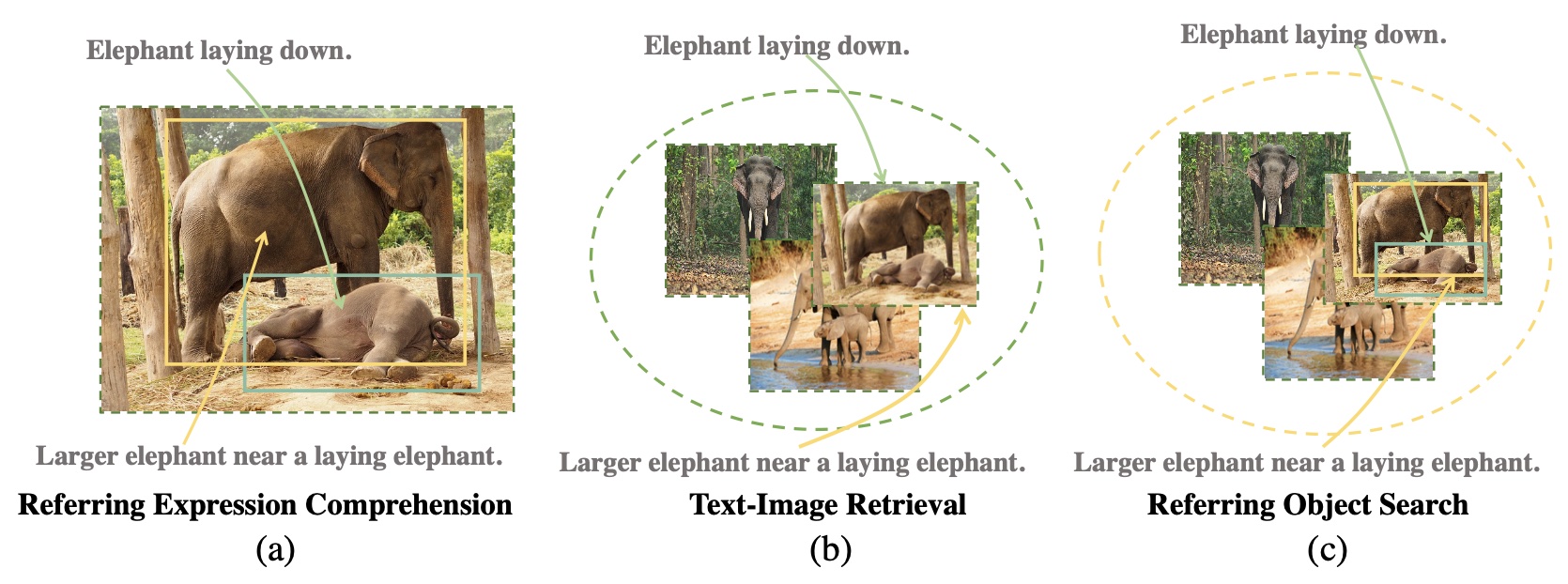}
\caption{This figure illustrates the difference between Referring Image Comprehension~(RIC), Text-Image Retrieval and ReSeeD. The text of RIC is to estimate a targeted object \textbf{in one image},while Text-Image Retrieval focus \textbf{on whole image searching in large database.} Our propose ReSeDis intends to \textbf{search target objects in all images in large database.}}
\label{fig:introduction_ror}
\end{figure*}

When searching for a specific object—such as ``a small yellow mug on the third shelf'' within a large image collection, existing methods typically fall into two categories. (1) Visual Grounding~\cite{CGAN-ACMMM2020, LAVT-CVPR2022, EFN-CVPR2021, BRINet-CVPR2020, BUSNet-CVPR2021, CRIS-CVPR2022, su2023language, hu2017modeling, zhang2018grounding, MAttNet-CVPR2018, liu2019improving, yang2019dynamic, hong2019learning, liu2019learning,zhou2021real, yang2019fast, liao2020real, yang2020improving, sun2021iterative, deng2021transvg, su2023language} focuses on localizing objects referred to by natural language expressions within a single image, typically by predicting bounding boxes or segmentation masks. However, these methods are developed under a closed-world assumption: the referred object is assumed to be present in every given image. As a result, even if the object does not exist in the image, the model is still forced to return a prediction, often yielding incorrect or random results. While effective in well-curated datasets, this assumption severely limits their utility in open-world scenarios where most images may not contain the target object, as in Figure~\ref{fig:introduction_ror}~(a). (2) Text-to-Image Retrieval~\cite{chung2014empirical, qu2020context, chen2021learning, zeng2022learning, huang2018bi, diao2021similarity} approaches the problem from the opposite direction. 
These methods rank images from a large collection based on their overall similarity to the input expression using vision-language models. While scalable and effective at retrieving semantically relevant images, they do not perform fine-grained localization and therefore cannot provide precise object-level information such as bounding boxes or segmentation masks, as in Figure~\ref{fig:introduction_ror}~(b).

While both methods have significantly driven advances in multimodal technology, their individual limitations become particularly problematic at large-scale scenarios: Visual Grounding methods, constrained by unrealistic assumptions, generate false positives when the object is not actually present, whereas Text-to-Image Retrieval, lacking fine-grained localization ability. Consequently, even minor inaccuracies inherent to both approaches can produce substantial false-positive results, limiting their practical usability in real-world fine grained visual search. To tackle these critical gaps, we introduce a novel task: Referring Search and Discovery (ReSeDis), as in Figure~\ref{fig:introduction_ror}~(c). ReSeDis uniquely integrates the strengths of large-scale image retrieval and precise localization. Given a textual query, it aims to efficiently search extensive image databases to identify all images containing the described object, and then accurately localize the object within each image using bounding boxes and segmentation masks. Thus, ReSeDis directly addresses two crucial, intertwined questions—"Does this image contain the object?" and "Where exactly is it located?".

One key challenge for ReSeDis is the lack of suitable datasets for evaluation. Most visual grounding benchmarks assume that each text refers to an object in a single image, which does not fit open-world settings where the object may appear in only a few images. This often leads to confusion when similar objects exist elsewhere in the dataset. Text-image retrieval datasets, on the other hand, only provide image-level labels and do not include bounding boxes or segmentations annotation. Their descriptions usually cover the whole image, not specific objects, making them unsuitable for fine-grained object search. To overcome this, we build a new benchmark dataset that ensures each description maps to a unique and well-defined set of object instances distributed across the entire image collection. This removes ambiguity and enables consistent and fair evaluation. To accurately assess model performance on the ReSeDis task, we also introduce a new evaluation metric. This metric is designed to reflect the dual demands of retrieval and localization. By capturing both aspects of the task, this metric helps determine how well models truly understand the relationship between text descriptions and object instances across a large and diverse dataset. Finally, we introduce a simple but informative zero-shot baseline using a pre-trained vision-language model. This baseline sets a reference point for future studies and highlights the gap between current capabilities and the demands of the ReSeDis task.

In summary, our contributions are:

\begin{itemize}
\item \textbf{Task:} We propose Referring Search and Discovery (ReSeDis), a novel multimodal task that bridges large-scale image retrieval and fine-grained object localization within a unified framework.

\item \textbf{Dataset:} To support this task, we build a new benchmark dataset that captures the complexity of real-world scenarios, where each textual query maps to well-annotated object instances distributed across a wide image collection.

\item \textbf{Metric:} We introduce a task-specific evaluation metric that jointly measures retrieval accuracy and localization precision.

\item \textbf{Baseline:} We provide a zero-shot baseline based on a frozen vision-language model, offering a practical starting point for future research.
\end{itemize}

\begin{figure*}[h]
    \centering
    \begin{subfigure}[b]{0.40\textwidth}
        \centering
        \raisebox{4mm}{ 
            \hspace{3mm} 
            \includegraphics[width=\textwidth]{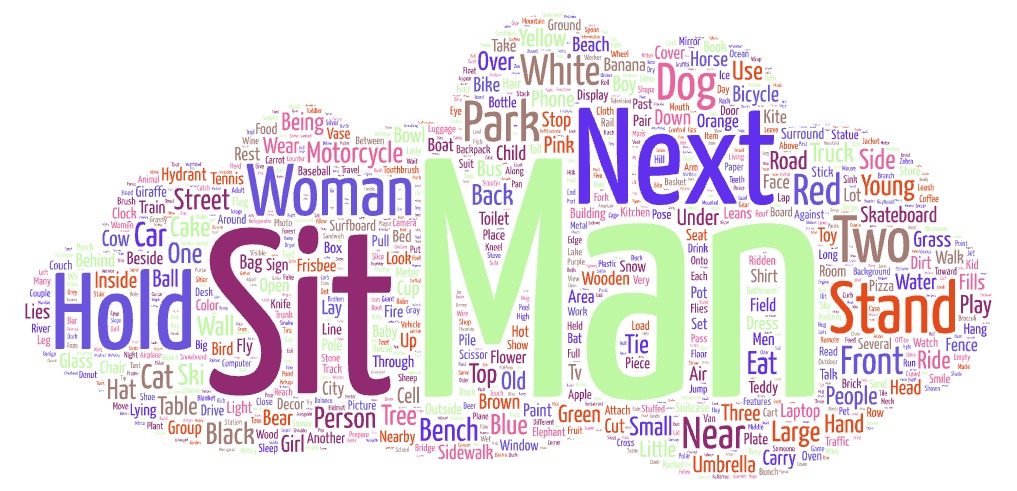}
        }
        \label{fig:category1}
    \end{subfigure}
    \hfill
    \begin{subfigure}[b]{0.55\textwidth}
        \centering
        \includegraphics[width=\textwidth]{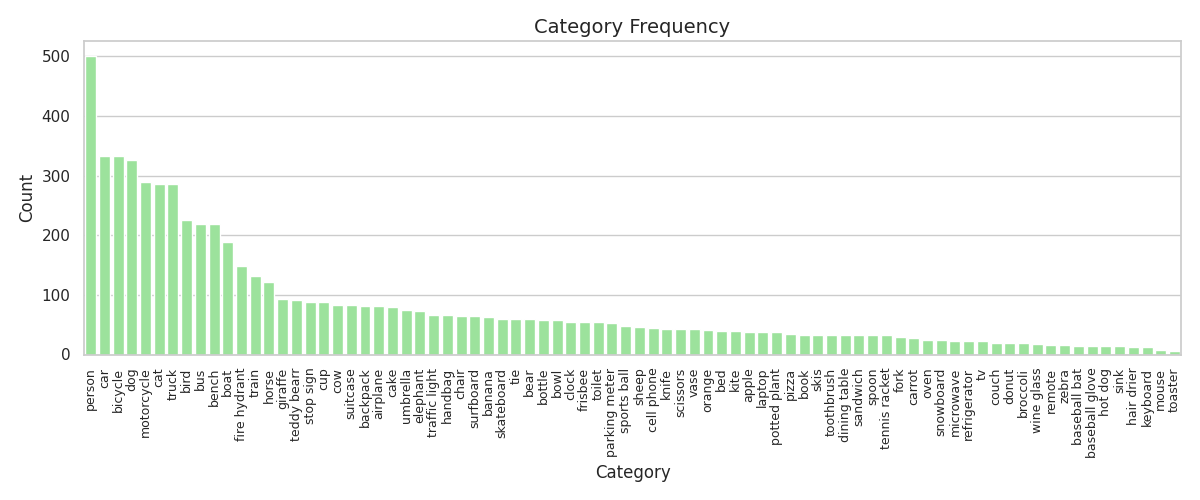}
        \label{fig:category2}
    \end{subfigure}
    \vspace{-25pt}
    \caption{Left: Word cloud generated from all expressions in the dataset, highlighting the most frequently occurring terms. Right: Distribution of categories across the dataset, with each bar representing the number of instances per category.}
    \label{fig:data_overview}
\end{figure*}



\begin{table*}[]
\centering
\begin{tabular}{lccccc}
\toprule
\textbf{Dataset} & \textbf{Task} & \textbf{Images} & \textbf{Expressions} & \textbf{Granularity} & \textbf{Open-world Search}  \\
\midrule
RefCOCO~\cite{yu2016modeling} & RIS\&RIC & 1500 & 10,834 & Segmentation & \xmark \\
RefCOCOg~\cite{mao2016generation} & RIS\&RIC & $\sim$2000 & 4896 & Segmentation & \xmark \\
GRES~\cite{liu2023gres} & RIS\&RIC & 1500 & 8163 & Segmentation & \xmark \\
\rowcolor{gray!10}
\textbf{ReSeDis (Ours)} & ReSeDis & \textbf{7088} & \textbf{9664} & Segmentation & \cmark  \\
\bottomrule
\end{tabular}
\caption{Comparison of validation split statistics across referring-related datasets. ReSeDis enables open-world search with globally consistent annotations.}
\label{tab:val_dataset_comparison}
\end{table*}

\section{Related Work}
Our proposed task ReSeDis is highly related to the task Referring Image Segmentation (RIS), Referring Image Comprehension (RIC), and text-image retrieval.

\subsection{RIS \& RIC}
Referring Image Segmentation (RIS) and Referring Image Comprehension (RIC) aim to localize objects in images based on natural language descriptions using segmentation masks~\cite{FRIST-ECCV2016, RMI-ICCV2017, DMN-ECCV2018} or bounding boxes~\cite{hu2016natural, MAttNet-CVPR2018}. Early RIS methods often employed one-step frameworks that combined text and image features, using techniques like memory units~\cite{RMI-ICCV2017}, attention mechanisms, and graph reasoning~\cite{CMPC-CVPR2020} to refine the predicted masks. Multi-step cascades, such as CGAN~\cite{CGAN-ACMMM2020, huang2023referring, huang2024loa}, progressively refine results by repeatedly fusing text and image features. RIC has two main approaches: two-stage methods~\cite{girshick2015fast, MAttNet-CVPR2018} that first propose regions and rank them based on text match, and one-stage methods~\cite{zhou2021real, deng2021transvg} that directly fuse text and image features, reducing error propagation and improving efficiency in bounding box prediction.

\subsection{Text-Image Retrieval}
Text-image retrieval helps systems find images based on textual queries and is used in search engines, digital asset management, and recommendation systems. It includes three methods: global matching, which uses CNNs for visual features and RNNs or GRUs for text features to calculate similarity~\cite{chung2014empirical}; regional matching, which focuses on aligning specific image regions with text for more precise retrieval~\cite{qu2020context, chen2021learning}; and multi-level matching, which integrates global and regional features, often using graph convolutional networks or attention networks for better accuracy~\cite{zeng2022learning, huang2018bi, diao2021similarity}. These strategies enhance text-image alignment, improving retrieval performance.

Our work connects to three well-known tasks but goes beyond each of them. Like Referring Image Segmentation (RIS) and Referring Image Comprehension (RIC), we link a sentence to the visual features of an object so we can point out exactly what the text is talking about. Like text-image retrieval, we start by searching a large image pool to find pictures that might contain that object. The key difference is what happens next: RIS and RIC assume the object is already in the given image, while retrieval methods return whole images and never show where the object is. In contrast, we first decide which images truly contain the object and then highlight its precise pixels, unifying large-scale search with fine-grained localization in a single pipeline.

\section{Dataset}
\subsection{Dataset Overview}
ReSeDis is a benchmark specifically designed to evaluate a model’s ability to perform large-scale referring object search in realistic, open-world settings. Unlike traditional referring tasks that assume the referred object is already present in a single image, ReSeDis asks a more challenging question: Given a natural language expression, can a model search through thousands of images, find the ones that truly contain the object, and accurately localize it at the pixel level? To evaluate this capability, we construct a validation-only dataset consisting of 7,088 images from the MS-COCO~\cite{lin2014microsoft} dataset and 9,664 manually written referring expressions. These expressions span all 80 MS-COCO categories and reflect real-world language, including object names, attributes (e.g., ``red'', ``tall''), actions (e.g., ``riding'', ``holding''), and spatial cues (e.g., ``on the left'', ``next to the dog''). Each sentence may refer to one object, multiple objects in a certain image inside image collections, meaning models must be able to perform both retrieval (which images are relevant) and localization (where the object is). We show some examples in Appendix.

The dataset follows a long-tailed category distribution, consistent with MS-COCO dataset. For example, the ``person'' category appears in nearly 3,500 expressions, while many other categories occur far less frequently. For visualization purposes, we clip the y-axis at 500 in our category distribution plot. To further illustrate the dataset's language diversity, we include a word cloud generated from all referring expressions, showing the rich variety of descriptors used—ranging from colors and shapes to actions and relative positions—making the benchmark closer to how humans naturally describe objects, as shown in Figre~\ref{fig:data_overview}.

We do not provide a fixed training set for ReSeDis because there are already several high-quality RIS/RIC datasets available, such as RefCOCO~\cite{yu2016modeling}, RefCOCOg~\cite{mao2016generation}, and GRES~\cite{liu2023gres}. While these datasets were not specifically designed for our task, they can still be effectively used during training, which is similar to real-world settings—where training images are often collected from the web—models must decide whether a given description refers to an object in the image (positive pair) or not (negative pair). This decision process is an important part of model training and reflects the open-world nature of our task. By leaving the choice of training data open, ReSeDis encourages flexible and realistic training strategies, while providing a standardized and challenging benchmark for evaluation.

\subsection{Dataset Comparison}
Table~\ref{tab:val_dataset_comparison} compares several widely used benchmarks in the referring vision-language field. Early datasets such as RefCOCO~\cite{yu2016modeling} and RefCOCOg~\cite{mao2016generation} are designed for intra-image grounding, where the referred object is always known to exist within the given image. These datasets have supported significant progress in both referring image comprehension (RIC) and referring image segmentation (RIS), but they operate under a closed-world assumption—they do not test whether a system can search for the object across multiple images, nor do they include truly negative cases.

GRES~\cite{liu2023gres} takes a step forward by introducing negative pairs, allowing models to judge whether a referred object is present in an image. This adds a layer of realism, but its setup remains limited to evaluating one query per image, meaning models still do not perform retrieval across an image collection. The task becomes a binary decision—is the object here or not—rather than an open search problem.

In contrast, ReSeDis is the first benchmark to fully support retrieval-based referring understanding in an open-world setting. Each referring expression in ReSeDis is uniquely linked to one or more objects in one image across a large image corpus. This setting not only retains the fine-grained mask-level annotations used in RIS, but also introduces a more realistic retrieval component: models must scan the entire dataset to find relevant images, then accurately localize the object. As shown in Table~\ref{tab:val_dataset_comparison}, ReSeDis bridges the gap between traditional referring tasks and real-world applications—such as web image search or visual surveillance—where the target object may appear anywhere.

\subsection{Dataset Preparation}
\begin{figure*}[h]
\centering
\includegraphics[width=0.95\linewidth]{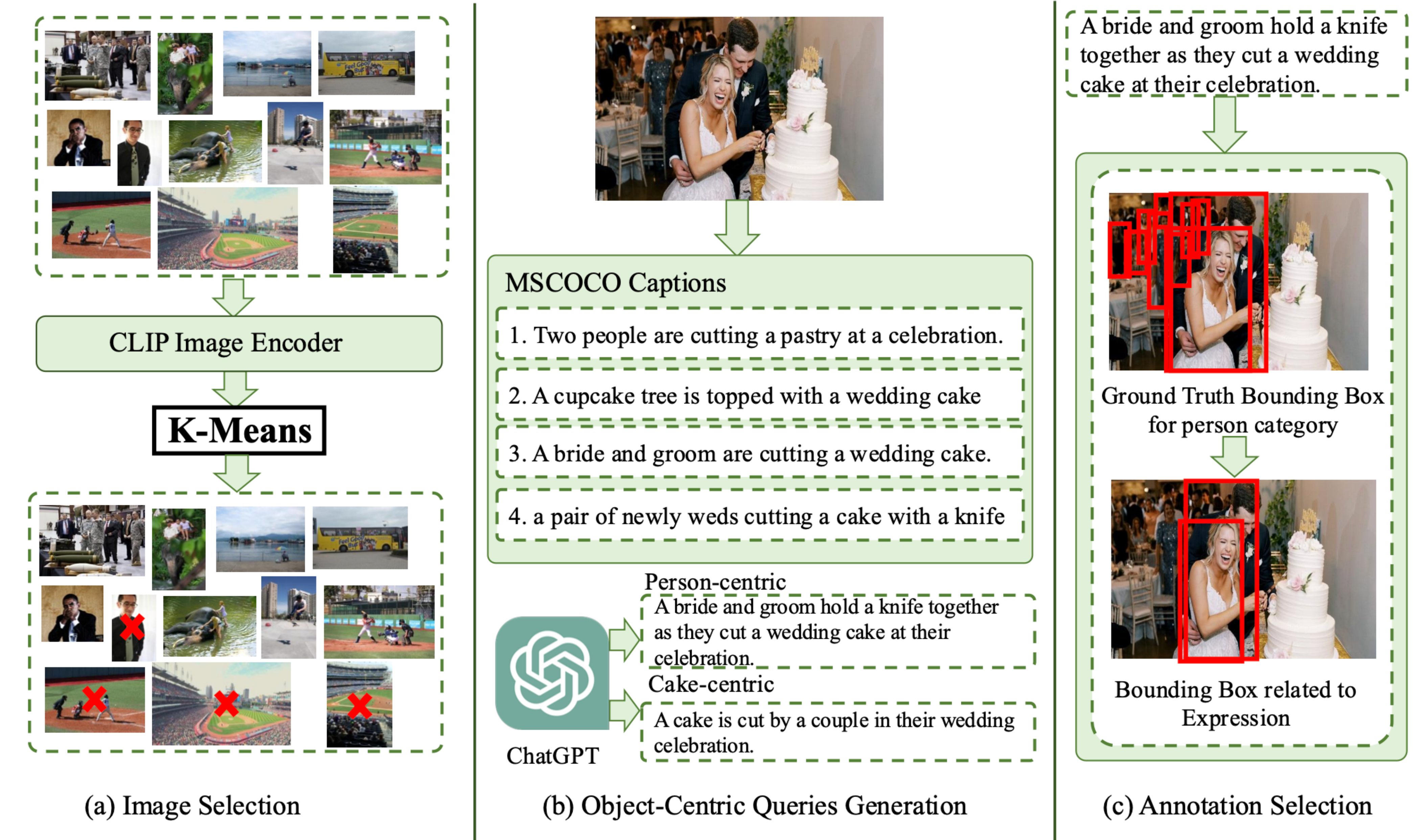}
\caption{The Overall pipeline for dataset construction.}
\label{fig:dp}
\end{figure*}

The motivation behind preparing our new dataset is to overcome the limitations of existing datasets for Referring Image Segmentation (RIS) and Referring Image Comprehension (RIC) tasks. Our goal is to create a resource that ensures precise and unique correspondence between text descriptions and objects. Specifically, our dataset aims to achieve the following targets:

\underline{Multiple Targets in One Image:} Each image in the dataset contains multiple targeted instances, where each target has a unique description. These descriptions focus exclusively on the specific target, detailing its attributes and the relationships between the target and other objects within the same image. Each target within an image is described individually, ensuring that different objects are distinctly identified and characterized.

\underline{Instance Correspondence:} Each description may refer to one or more instances within an image. Accurate identification and annotation of these instances, using bounding boxes and segmentation masks, are crucial to ensure proper object localization and alignment with the given descriptions.

\underline{Unique Correspondence:} Each description corresponds to only one image in the entire dataset. While typically a description may refer to multiple images, balancing the number of images is essential for the recall accuracy calculation. Thus, for the purpose of this calculation, each description is linked to a unique image.

To realize these goals, MSCOCO~\cite{lin2014microsoft} is chosen as the foundational data source due to its extensive collection of images captured in diverse environments, detailed bounding-box and segmentation mask annotations, and rich caption annotations. Then, we undertook the following steps, as shown in Figure~\ref{fig:dp}:

\textbf{Image Selectin} To remove heavily overlapping images in the MSCOCO validation dataset and reduce the workload for further labeling, we perform image selection at the initial stage. We begin by using the CLIP Image Encoder to extract features from all images in the MSCOCO validation dataset. These features help identify visual similarities among the images. Next, we apply the K-Means algorithm to cluster these images into 10 groups initially. We then refine each group by reapplying K-Means two more times in the same manner, resulting in 1,000 clusters. Afterward, we manually review these 1,000 clusters and retain visually similar images that can be differentiated using text descriptions. By focusing on the textual distinctions, we ensure diversity in the dataset while maintaining visual similarity. This process results in a final selection of around 10,000 distinct images, ensuring a variety of unique visual content with clear text-based distinctions, as in Figure~\ref{fig:dp}~(a).

\textbf{Object-Centric Queries Generation}  After selecting images from the MSCOCO dataset, we enhance the annotations by using ChatGPT to generate detailed, object-centric descriptions for each image. This process involves creating prompts that include both the object's category and contextual information from existing MSCOCO annotations. For example, consider an image from the dataset showing a wedding ceremony, where MSCOCO includes categories such as "person" and "cake." For this image, we would generate two distinct descriptions using ChatGPT: one focused on the people involved, such as "A bride and groom hold a knife together as they cut a wedding cake at their celebration," and another focusing on the cake, such as "A cake is being cut by a couple." This dual-description approach ensures that each significant element in the image is described, enhancing the dataset with richer, more nuanced annotations that capture both the primary subjects and their interactions within the scene, as in Figure~\ref{fig:dp}~(b). To achieve unique correspondence, the image inside the same group will be annotated carefully by attribute and their environment manually. 

\textbf{Annotation Selection} In the previous step, we generated a detailed, person-centric expression: "A bride and groom hold a knife together as they cut a wedding cake at their celebration." Since MSCOCO only provides bounding box and segmentation mask annotations for persons, the next step is to carefully select the corresponding bounding box and segmentation mask that accurately represent the bride and groom in the description. This involves manually removing any extraneous annotations that do not pertain to the described individuals. By doing this, we ensure that the visual data matches the generated text precisely, as in Figure~\ref{fig:dp}~(c).

\section{Evaluation Metric}
In the Referring Search and Discovery (ReSeDis) task, we evaluate model performance in two key areas:

\begin{itemize}
    \item \textbf{Recall Accuracy of Target Images}: This metric measures how well the model can correctly identify images containing the referred object. High recall accuracy is essential as it shows the model’s ability to effectively narrow down the search space to the relevant images.
    
    \item \textbf{Object Localization Accuracy}: After identifying the target images, this metric assesses how accurately the model can locate the correct object in all images. Specifically, we use Precision at 50\% overlap threshold (Pr@50), which calculates the precision of object detection by considering a prediction correct if the detected object has at least 50\% overlap with the ground truth bounding box.
\end{itemize}

\section{Method}
\begin{figure*}[ht]
\centering
\includegraphics[width=1.00\linewidth, height=0.30\linewidth]{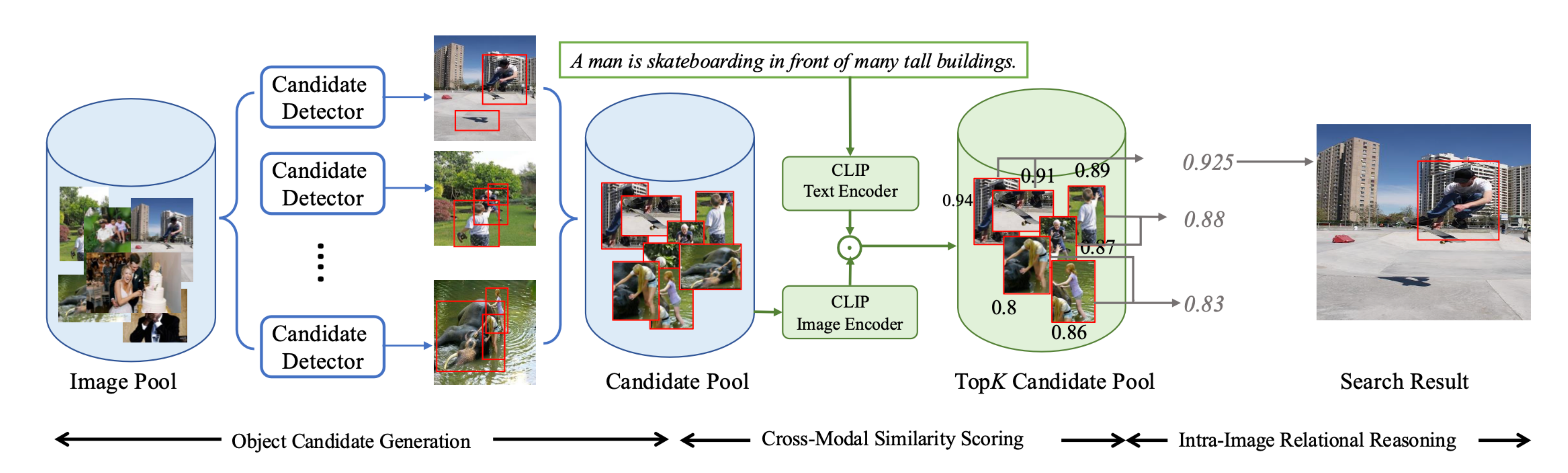}
\caption{The illustration of our proposed baseline. We use Yolov8 as candidate detector.}
\label{fig:baseline}
\end{figure*}

Our goal is to find specific objects in a large image dataset based on a natural language description. To do this, we propose a simple two-stage framework that includes object-level localization and matching in all images, and fine-grained reasoning over object attributes and inter-object relationships. The method contains three phrases: Object Candidate Generation, Cross-Modal Similarity Scoring, Intra-Image Relational Reasoning, as in Figure~\ref{fig:baseline}.

\subsection{Object Candidate Generation}
Given a natural language expression $L = \{l_i\}{i=1}^l$ and an image set $I = \{I_1, I_2, ..., I_t\}$, where $l$ is the length of the expression and $t$ is the total number of images, the first step is to reduce the search space by identifying a manageable set of object candidates in each image. Exhaustively comparing the expression against every possible region in every image would be computationally prohibitive. To address this, we employ YOLOv8~\cite{Jocher_Ultralytics_YOLO_2023}, a state-of-the-art object detection model known for its speed and accuracy. For each image $I_t$, YOLOv8 generates a set of object candidates $\{I_t^k\}_{k=1}^{n}$, where $n$ is the number of detected regions in image $I_t$, as in Equation \ref{eq:candidate}. These proposals serve as the candidate objects that are potentially referred to by the expression and are passed to subsequent modules for cross-modal matching and reasoning.

\begin{equation}
\{I_t^k\}_{k=1}^{n} = YOLO_{v8}(I_t)
\label{eq:candidate}
\end{equation}

\subsection{Cross-Modal Similarity Scoring}
Then, each detected object candidate is resized and fed into the CLIP~\cite{radford2021clip} image encoder to extract its visual features. In parallel, the natural language expression $L$ is encoded using the CLIP text encoder to obtain its textual representation. CLIP projects both modalities into a shared embedding space, enabling direct comparison. The cosine similarity score $S$ between the text and each object proposal is calculated as:

\begin{equation}
S_t^f = \frac{L \cdot I_t^f}{|L|\cdot|I_t^f|}
\label{eq:rosd_cosine}
\end{equation}

This similarity score $S_t^f$ reflects how well each candidate region individually aligns with the input expression. However, this step considers each object proposal in isolation and lacks global contextual information, such as the relationships among objects or the overall scene structure. As a result, it may not be sufficient to determine whether a highly similar candidate is indeed the correct referent. Therefore, we introduce a subsequent reasoning module to incorporate inter-object relationships within each image, enabling more accurate and context-aware selection.

\subsection{Intra-Image Relational Reasoning}
To make our matching results more accurate, we also look at how objects relate to each other within the same image. First, we pick the top-$K$ object proposals that have the highest similarity scores $S_{tf}$ with the text description. These are the most likely matches based on how well each region alone fits the expression. Then, for each image that contains one or more of these top-$K$ proposals, we calculate the average of their similarity scores, as in Equation~\ref{eq:reasoning}. This gives us a single score for the whole image, which reflects not only the quality of the individual matches but also how well the group of objects in the image together match the description. After that, we rank all images using their average scores. The images with the highest scores are selected as the best matches, and the top object proposals from these images are used as the final search results.

\begin{equation}
S_{t} = \frac{1}{K} \sum_{k=1}^{K} S_{t}^{k}
\label{eq:reasoning}
\end{equation}

By combining both region-level matching and image-level reasoning, our method can better understand the full meaning of the expression—not just which objects are mentioned, but also how they relate to each other.

\subsection{Results}
\begin{table}[ht]
\centering
\begin{tabular}{c|cc|cc}
\toprule
\multirow{2}{*}{Setting} & \multicolumn{2}{c|}{Top-1} & \multicolumn{2}{c}{Top-5} \\
\cmidrule{2-5}
& Recall & Pr@50 & Recall & Pr@50 \\
\midrule
YOLOv8 Detection & 0.2122 & 0.03257 & 0.4206 & 0.06152 \\
Ground-Truth Boxes & 0.2333 & 0.02717 & 0.4244 & 0.04904 \\
\bottomrule
\end{tabular}
\caption{Top-k, k==100 retrieval performance: comparison between YOLOv8 proposals and ground-truth boxes. We report Top-1 and Top-5 image recall and corresponding localization precision (Pr@50).}
\label{tab:retrieval_top100}
\end{table}

We analyze retrieval and localization performance using CLIP ViT-B/32 under the Top-100 candidate setting. As shown in Table~\ref{tab:retrieval_top100}, when using proposals generated by YOLOv8, the model achieves a Top-1 image recall of 21.22\% with a localization precision (Pr@50) of 3.57\%. Top-5 recall improves significantly to 42.06\%, along with a higher Pr@50 of 6.15\%, indicating that while the model often includes the correct image within the top few ranks, its top prediction is still prone to errors.

When using ground-truth bounding boxes as proposals, performance improves slightly in Top-1 recall (23.33\%) and Top-5 recall (42.44\%), but Pr@50 actually drops. This suggests that while using perfect proposals helps with recall, it does not necessarily improve localization precision under our evaluation metric. One possible explanation is that YOLOv8 proposals are better tuned to object boundaries, which affects the intersection-over-union (IoU) based precision calculation. Overall, these results highlight both the strengths and limitations of CLIP ViT-B/32: it shows promising retrieval ability, especially within the top 5 ranks, but still struggles with precise localization at the object level. More results can be found in Appendix.

\section{Real-World Applications for the Task ReSeDis}
The ReSeDis task is designed to help systems understand and connect natural language with visual content across large and unstructured collections of images. Unlike traditional tasks that focus on one image at a time, ReSeDis allows a model to search through many images and find exactly where the described object appears. This opens the door to many real-world uses:

\textbf{(1) Smart Surveillance Systems:} In busy places like airports, train stations, or public events, security staff often need to find someone quickly based on a description. For example, they might say, ``a man in a red jacket carrying a black backpack near the entrance.'' ReSeDis makes it possible to search all available camera feeds and highlight the right person without checking each frame by hand. This could help locate lost individuals, track suspects, or manage crowds more efficiently.

\textbf{(2) Agriculture Monitoring} Farmers and agricultural experts can use drones to capture images of fields and describe what they are looking for, such as ``wilted plants with brown edges near the irrigation pipe.'' A system trained for ReSeDis could automatically scan all drone images, find where the problem is happening, and mark the exact locations. This helps with early detection of issues, reduces pesticide use, and supports precision farming to improve crop health.

\textbf{(3) Medical Image Search and Diagnosis:} In hospitals, doctors often need to look up similar cases in large databases of medical images. With ReSeDis, they could search using descriptions like ``a round lesion with uneven edges in the lower left lung.'' The system could return relevant images, showing not only the correct scans but also highlighting the areas of concern. This can support diagnosis, training, and treatment planning—especially when combined with patient records.

\textbf{(4) Environmental Change Detection:} Researchers studying climate change or ecosystems often rely on satellite and aerial images. With ReSeDis, they can issue natural language queries such as ``coral reefs next to mangrove forests'' or ``deforested patches near rivers.'' The system can find matching images and highlight the regions of interest, helping scientists monitor changes over time, track damage, and provide data to inform environmental policy.

\textbf{(5) Online Shopping:} Shoppers often know what they want but can’t describe it with exact keywords. For example, a user might say: ``a beige crossbody bag with a gold chain, worn by a woman in a street-style photo.'' A ReSeDis-based system could search across product catalogs, influencer photos, or social media images, and not only retrieve relevant results, but also highlight the specific object being referred to in each image. This enables a more natural and intuitive shopping experience, especially in cases where visual context—like how the product is worn or styled—is essential to the user’s intent.

\bibliographystyle{ACM-Reference-Format}
\bibliography{acmart}

\appendix
\newpage

\end{document}